\pdfoutput=1
\documentclass[11pt]{article}

\usepackage{ACL2023}

\usepackage{times}
\usepackage{latexsym}
\usepackage{dirtytalk}

\usepackage{cjhebrew} % http://mirrors.ibiblio.org/CTAN/language/hebrew/cjhebrew/cjhebrew.pdf

\usepackage[T1]{fontenc}
\usepackage[utf8]{inputenc}

\usepackage{microtype}

\usepackage{inconsolata}

\usepackage{tikz}
\usepackage{graphicx}
\usetikzlibrary{arrows.meta, positioning}
\usepackage{amsmath, amssymb}

\usepackage{booktabs}
\usepackage{xcolor}
\usepackage{colortbl}
\usepackage{multirow}
\usepackage{algorithm}
\usepackage{algpseudocode}

\newcommand{\spl}{\textsc{Splinter}}
\newcommand{\baseline}{Vanilla}
\newcommand{\bpe}{BPE}
\newcommand{\unigram}{UnigramLM}
\newcommand{\spls}{Splinters}

\title{Splintering Nonconcatenative Languages for Better Tokenization}

\author{
        Bar Gazit$^{\beta\lambda}$ \quad \quad 
        Shaltiel Shmidman$^{\delta}$ \quad \quad 
        Avi Shmidman$^{\chi\delta}$ \quad \quad 
        Yuval Pinter$^{\beta\lambda}$ \\
        \begin{tabular}{cc}
        $^{\beta}$Department of Computer Science, $^{\lambda}$Data Science Research Center 
                & $^{\chi}$Bar-Ilan University \\
        Ben-Gurion University of the Negev 
                & Ramat Gan, Israel \\ 
                Be'er Sheva, Israel & $^{\delta}$DICTA, Jerusalem, Israel \\ 
        \end{tabular}
        \\
        \texttt{\small 
            \{bargazi@post,uvp@cs\}.bgu.ac.il  \quad shaltieltzion@gmail.com \quad avi.shmidman@biu.ac.il}
    }

\begin{document}
\maketitle

\begin{abstract}
Common subword tokenization algorithms like BPE and UnigramLM assume that text can be split into meaningful units by concatenative measures alone.
This is not true for languages such as Hebrew and Arabic, where morphology is encoded in root-template patterns, or Malay and Georgian, where split affixes are common.
We present \spl{}, a pre-processing step which rearranges text into a linear form that better represents such nonconcatenative morphologies, enabling meaningful contiguous segments to be found by the tokenizer.
We demonstrate \spl{}'s merit using both intrinsic measures evaluating token vocabularies in Hebrew, Arabic, and Malay; as well as on downstream tasks using BERT-architecture models trained for Hebrew.
\end{abstract}

\section{Introduction}
\label{sec:introduction}

Large language models (LLMs) have become pivotal in natural language processing (NLP), offering extensive utility across diverse applications.
Central to constructing an LLM is producing basic input units from the text sequence, for which subword tokenization is still the standard approach, using methods such as byte-pair encoding (BPE)~\cite{sennrich-etal-2016-neural}, WordPiece~\cite{schuster2012japanese}, and UnigramLM~\cite{kudo-2018-subword}.
However, subword tokenizers exhibit diminished effectiveness in nonconcatenative languages (NCLs) such as Hebrew and Arabic~\cite{klein-tsarfaty-2020-getting}.
While tokenizers assume linear segmentation of words, NCLs' units of meaning are typically intertwined within words, and \textbf{cannot be separated linearly}, as the root letters are not adjacent~\cite{khaliq-carroll-2013-induction}.
A Hebrew example is provided in \autoref{tab:2-sentences-example}.
in the Hebrew word \<l`bwd> \emph{la'avod} `to work', the root letters are `\<`>', `\<b>', and `\<d>', placed in an infinitive morphological template manifested by the locations of `\<l>' and `\<w>'.
This characteristic forces linear tokenizers to split words into morphologically-incoherent tokens, losing the downstream models' ability to generalize across various forms of the same lemma, and eventually reducing model performance when applied to a large variety of tasks such as text generation and translation~\cite{keren2022breaking,levi2024truly,shmidman-etal-2024-mrl}.

\begin{table}
    \centering
    \begin{tabular}{lc} \toprule
        \multirow{2}{*}{(a)} & .\cjRL{`bdty `l hm.sgt} \\
        & `I worked on the presentation.' \\
        \multirow{2}{*}{(b)} & .\cjRL{h`bwdh hq/sh h/stlmh} \\
        & `The hard work paid off.' \\ \bottomrule
    \end{tabular}
    \caption{Examples of Hebrew text (read from right to left) exemplifying its nonconcatenative morphology. The root \cjRL{`bd} `work' appears in both sentences, but in (a) it comprises a linear segment of the text whereas in (b) it is broken by the templatic character \cjRL{w}.}
    \label{tab:2-sentences-example}
\end{table}

We present \spl{}, a statistical algorithm for \textbf{linearizing} NCL text through rearranging the text sequence by iteratively pruning characters from words, with the intent of isolating characters representing template forms.
The manipulated text can then be input into any ordinary linear tokenizer for processing as usual, adapting the NCL data into the morphologically-concatenative input BPE and its like expect.
We show that vocabularies and models trained over \spl{}-processed text outperform those starting from raw NCL text on both intrinsic and extrinsic measures in Hebrew, Arabic, and Malay.\footnote{\url{https://github.com/MeLeLBGU/Splintering}.}

\section{Tokenizing with \spls}

\begin{figure*}[ht!]
\centering
\scalebox{0.85}{
\begin{tikzpicture}[
    every node/.style={font=\sffamily, align=center},
    arrow/.style={-{Stealth}, thick},
    highlight/.style={draw=blue, thick, rounded corners, inner sep=3pt},
    textstyle/.style={yshift=10pt, text width=50pt, align=center, text height=5ex},
    ]

% REGULAR FLOW %%%%%%%%%%%%%%%%%%%%%%%%%%%%%%%%%%%%%%%%%%
    % Title
    \node[font=\large\bfseries, align=center] at (-2, 2) {Standard Flow:};
    
    % Original Word
    \node[draw, minimum height=2.5em] (input_r) {
        \tikz \node[draw, fill=gray!20, minimum height=2em] (input_r_inner1) 
            {\textbf{‘\<d>’}};
        \tikz \node[draw, fill=gray!20, minimum height=2em] (input_r_inner2) 
            {\textbf{‘\<w>’}};
        \tikz \node[draw, fill=gray!20, minimum height=2em] (input_r_inner3) 
            {\textbf{‘\<b>’}};
        \tikz \node[draw, fill=gray!20, minimum height=2em] (input_r_inner4) 
            {\textbf{‘\<`>’}};
        \tikz \node[draw, fill=gray!20, minimum height=2em] (input_r_inner5) 
            {\textbf{‘\<l>’}};
    };
    \node[below=0.5em of input_r] {Original text};
    
    % incoming arrow into (input)
    \draw[arrow, line width=1.5pt] (-3.3,0) -- (input_r.west)
        node[midway, above, textstyle, text depth=1em] {Input};
        
    % Tokenizer
    \node[draw, minimum height=2.5em, right=of input_r] (tokenizer_step_r) {
        \tikz \node[draw, fill=gray!20, minimum width=4em, minimum height=2em] (inner1_r) 
            {\textbf{‘\<bwd>’}};
        \tikz \node[draw, fill=gray!20, minimum width=4em, minimum height=2em] (inner2_r) 
            {\textbf{‘\<l`>’}};
    };
    \node[below=0.5em of tokenizer_step_r] {Segmented text};
    \draw[arrow, line width=1.5pt] (input_r) -- (tokenizer_step_r) 
        node[midway, above, textstyle, text depth=1em] {Tokenizer};
    
    % LM
    \node[draw, minimum width=2.5em, minimum height=2.5em, right=of tokenizer_step_r] (lm_step) {\textbf{LM}};
    \draw[arrow, line width=1.5pt] (tokenizer_step_r) -- (lm_step) 
        node[midway, above, textstyle, text depth=1em] {Embedding};

% OUR FLOW %%%%%%%%%%%%%%%%%%%%%%%%%%%%%%%%%%%%%%%%%%
    % Title
    \node[font=\large\bfseries, align=center] at (-2, -2) {Splinter Flow:};
    
    % Original Word
    \node[draw, minimum height=2.5em] at (0, -4) (input) {
        \tikz \node[draw, fill=gray!20, minimum height=2em] (input_inner1) 
            {\textbf{‘\<d>’}};
        \tikz \node[draw, fill=gray!20, minimum height=2em] (input_inner2) 
            {\textbf{‘\<w>’}};
        \tikz \node[draw, fill=gray!20, minimum height=2em] (input_inner3) 
            {\textbf{‘\<b>’}};
        \tikz \node[draw, fill=gray!20, minimum height=2em] (input_inner4) 
            {\textbf{‘\<`>’}};
        \tikz \node[draw, fill=gray!20, minimum height=2em] (input_inner5) 
            {\textbf{‘\<l>’}};
    };
    \node[below=0.5em of input] {Original text};
    
    % incoming arrow into (input)
    \draw[arrow, line width=1.5pt] (-3.3,-4) -- (input.west)
        node[midway, above, textstyle, text depth=1em] {Input};
    
    % Splintered
    \node[draw, minimum height=2.5em, right=of input] (splinter_step) {
        \tikz \node[draw, fill=gray!20, minimum height=2em] (input_r_inner5) 
            {\textbf{‘3:\<w>’}};
        \tikz \node[draw, fill=gray!20, minimum height=2em] (input_r_inner4) 
            {\textbf{‘0:\<l>’}};
        \tikz \node[draw, fill=gray!20, minimum height=2em] (input_r_inner1) 
            {\textbf{‘\<d>’}};
        \tikz \node[draw, fill=gray!20, minimum height=2em] (input_r_inner2) 
            {\textbf{‘\<b>’}};
        \tikz \node[draw, fill=gray!20, minimum height=2em] (input_r_inner3) 
            {\textbf{‘\<`>’}};
    };
    
    \node[below=0.5em of splinter_step] {Splintered text};
    \draw[arrow, line width=1.5pt] (input) -- (splinter_step) 
        node[midway, above, textstyle, text depth=1em] {\spl};
    
    % Tokenizer
    \node[draw, minimum height=2.5em, right=of splinter_step] (tokenizer_step) {
        \tikz \node[draw, fill=gray!20, minimum width=4em, minimum height=2em] (inner2) {\textbf{‘3:\<w>\,0:\<l>’}};
        \tikz \node[draw, fill=gray!20, minimum width=4em, minimum height=2em] (inner1) {\textbf{‘\<`bd>’}};
    };
    \node[below=0.5em of tokenizer_step] {Segmented text};
    \draw[arrow, line width=1.5pt] (splinter_step) -- (tokenizer_step) 
        node[midway, above, textstyle, text depth=1em] {Tokenizer};
    
    % LM
    \node[draw, minimum width=2.5em, minimum height=2.5em, right=of tokenizer_step] (lm_step) {\textbf{LM}};
    \draw[arrow, line width=1.5pt] (tokenizer_step) -- (lm_step) 
        node[midway, above, textstyle, text depth=1em] {Embedding};

\end{tikzpicture}
}
\caption{Overview of a Hebrew language model pipeline using the example `to work': standard flow vs. incorporating \spl. This figure is a real example for the differences in tokenization using a \bpe\ tokenizer with a vocab size of 2000. Gray boxes are ordered from right to left.}
\label{fig:splintering}
\end{figure*}
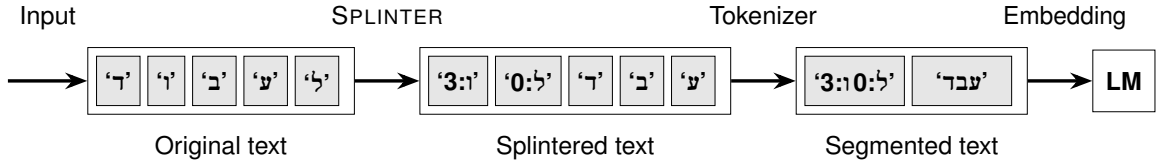

When designing our approach, our goal was to create a relinearized sequence for words in NCL languages, while adhering to several constraints.
One constraint is that the transformation must be reversible, ensuring that the new representation can always be converted back to the original text.
In addition, we aimed to develop a tool that would integrate smoothly with existing tokenizers, ensuring seamless adoption without requiring modifications in their implementation.
Additionally, the method should be applied only to the intended languages, without affecting the entire text. 
We also considered that the method should be adaptable to two distinct use cases: (1) models trained primarily in an NCL language, where the majority of the vocabulary belongs to that language (e.g.,~\texttt{DictaBERT}~\citep{shmidman2023dictabert} has a large vocabulary size of 128K); and (2) large-scale multilingual LLMs, where most tokens are allocated to English and only a small portion is left for the NCL language (e.g., GPT-4o, which has approximately 2.3K tokens allocated for Hebrew).
Thus, the method should be effective across different vocabulary sizes.

We develop \spl{} as a pre-tokenization step designed to address the challenges of subword tokenization in NCLs. 
The core idea of the algorithm is that in certain NCLs, many words are formed by embedding root letters into specific morphological templates. 
For instance, the Hebrew word \<l`bwd> is derived from the root \<`bd> placed in the template \<l\_\_w\_>.
Since there are far fewer templates than roots, template letters tend to appear in specific positions within a word more consistently than root letters do (e.g.,~for \<l`bwd>: \say{0:\<l>},\say{3:\<w>}).
Empirically, we observed that in Hebrew and Arabic, when a word is longer than 3 characters, there is always at least one deletion that, when applied, transitions the word to a different, existing template \textbf{while preserving the same root}.\footnote{However, not every letter's deletion will necessarily result in a valid template. For example, removing \<d> from \<l`bwd> would produce the non-word \<l`bw>.}
For instance, the word \<l`bwd> turns into \<l`bd> when the template letter \<w> is removed.
By repeating this process iteratively, the word eventually reduces to only its root letters.
This method can be seen as a way to isolate the root letters from the template.

In this method, NCL words are transformed into a sequence of single-letter reductions, with the goal of rearranging them to better align with existing subword tokenizers.
This is achieved by expanding the language's alphabet to include not only its original letters but also single-letter reductions, where each reduction consists of a letter paired with the index from which it was removed.\footnote{In practice, we found that using \emph{negative indices} for the latter half of a word's characters both aligns well with the suffixing nature of certain morphemes and reduces the resulting alphabet by 15\%. Not shown in our examples for simplicity. Further low-level details are provided in \autoref{sec:alphabet-encoding-implementation-details}.}
From another perspective, this method explores the impact of expanding a language's alphabet to enhance word representation.
We illustrate the high-level application of \spl{} into the NLP pipeline in \autoref{fig:splintering}.

\subsection{\spl{} Reduction Map Creation}
As mentioned above, the \spl\ pre-tokenization step processes a word by iteratively applying single-letter \emph{reductions}.
To achieve this, we must determine which reduction to perform at each iteration. 
We propose an algorithm that analyzes a given corpus and generates a mapping, where the keys represent word lengths and the values are ordered lists of reductions, ranked from most to least frequent.

\paragraph{Converting the Corpus to Unigram Frequencies}
The algorithm begins by processing a corpus in the target language, which undergoes several preprocessing steps. First, all diacritics are removed from the text (for Hebrew only). 
The text is then split into words using the following regex pattern:
\begin{verbatim}
\.|\s|\n|-|,|:|"|\(|\) 
\end{verbatim}
Next, words that appear fewer than 10 times in the corpus are discarded, along with any words containing letters from other languages.
Additionally, we normalized final and non-final Hebrew letters to maintain consistency in root-based word connections. 
Specifically, all final letters were replaced with their non-final forms, and vice versa when applicable (i.e.,~in cases where a non-final form occurs in the final position of a word, mostly in borrowed words like \<q.t/swp|> `ketchup').
This adjustment helps preserve morphological relationships, such as between \<hwlK> \emph{holex} `(he) is walking' and \<hwlkyM> \emph{holxim} `(they) are walking', both derived from the root \<hlk|> `walk'.
After this transformation, these words become \<hwlk|> and \<hwlkym|>, respectively, both clearly retaining the root \<hlk|>. 
The reverse transformation ensures that distinctions between different word groups are still maintained, keeping the original text recoverable.\footnote{Arabic also has final and non-final letter forms, but their selection occurs deterministically based on context. 
From a Unicode perspective, both forms share the same underlying character, eliminating the need for manual conversion.}
Finally, each remaining word is assigned its frequency count in the corpus, resulting in a \textbf{unigram dictionary} mapping words to their respective frequencies. 

\paragraph{Scoring the Reductions}
With the unigram dictionary constructed, we group the words by their length and iterate over them.
For each word $w_i$ of length $k\geq 4$,\footnote{We examine words with at least four characters since most Semitic roots contain three characters.} we evaluate all possible single-character reductions $\{r_0, r_1, \dots, r_{k-1}\}$, assigning each reduction a score equal to the frequency of the resulting word $w_i^{r_j}$ in the corpus, giving lower scores to rare words, and zero score if the resulting word is not in the dictionary.
For example, given the word $w_j$=\<lymwd> \emph{limud} `studying', we examine these possible reductions: $r_0$=\say{0:\<l>}, $r_1$=\say{1:\<y>}, $r_2$=\say{2:\<m|>}, $r_3$=\say{3:\<w>}, $r_4$=\say{4:\<d>}, which produce the following respective words:  $w_j^{r_0}$=\<ymwd>, $w_j^{r_1}$=\<lmwd>, $w_j^{r_2}$=\<lywd>, $w_j^{r_3}$=\<lymd>, $w_j^{r_4}$=\<lymw>, and score each of the reductions the frequency of the resulting word.
The score is summed for each word length $k$, per each reduction $r$, so as we iterate through the words, we build a map that tracks, for each word length, which reductions produced valid words, along with their scores. 
The resulting map is then sorted so that reductions are ranked from most to least frequent.

Following this step, the map is used as a starting point for a second iteration over the corpus. 
For each word of length $k$, we attempt to apply reductions from the map, traversing it by descending frequency score.
The first successful reduction that produces an existing word is recorded in a new frequency counter, using the same scoring method as in the first iteration, while all other possible reductions for that word are ignored.
This step ensures that the most frequent reduction is prioritized, aligning with our goal of removing template letters first, as they tend to appear more frequently in consistent locations than individual root letters.
Pseudocode for the full map creation algorithm is available in \autoref{app:algo}.

\begin{figure}
    \centering
    \includegraphics[width=0.45\textwidth]{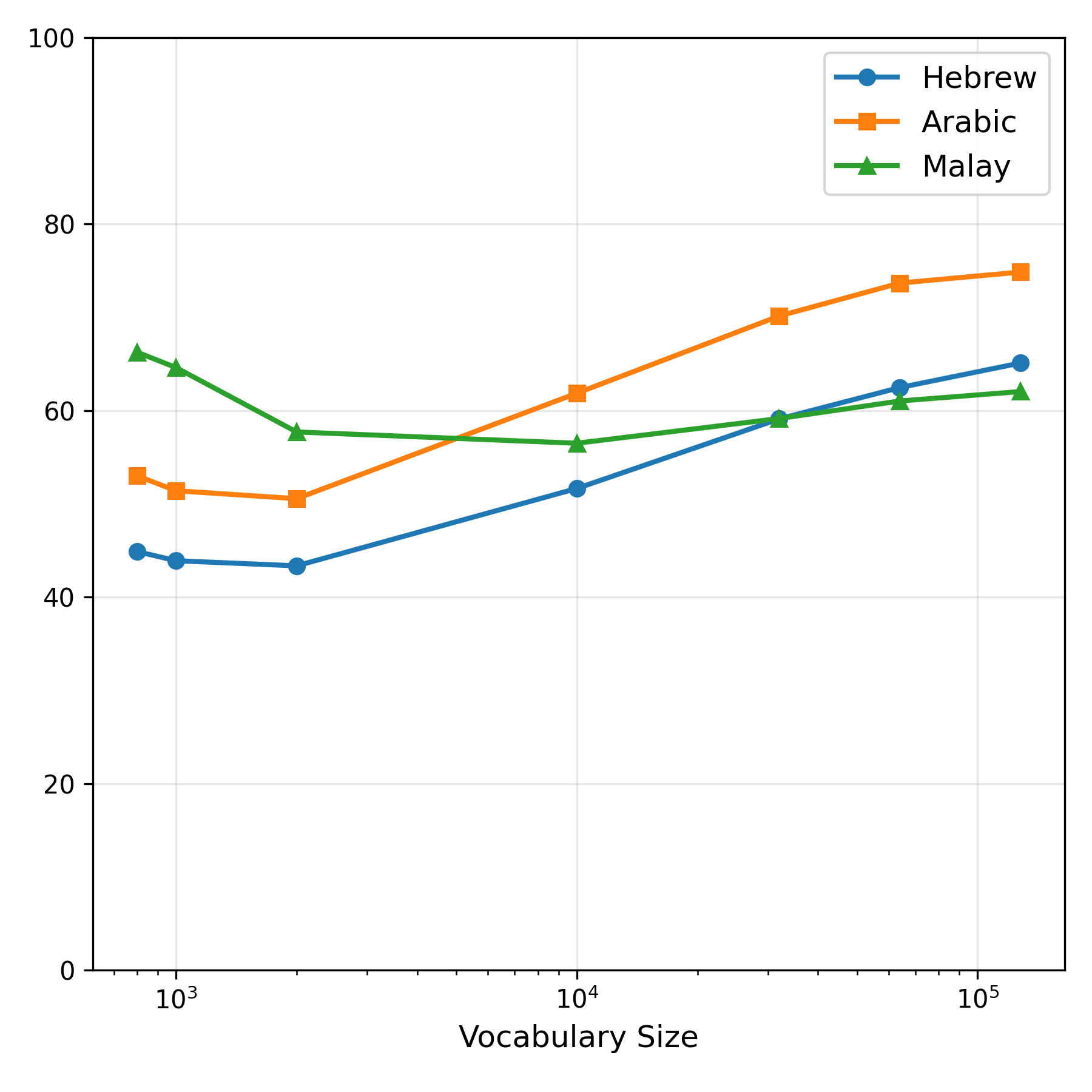}
    \caption{Intersection percent between \baseline\ \bpe\ and \bpe\ + \spl\ by vocabulary size for Hebrew, Arabic, and Malay. Vocabulary size is presented on a logarithmic scale.}
    \label{fig:intersection}
\end{figure}

\begin{table*}
    \centering
    \small
    \begin{tabular}{llrrrrrr}
    \toprule
    
    Tokenizer & Type & Cognitive    & Rényi      & Tokens   & 4+ token & 1-char & Distinct \\
              &      & plausibility & efficiency & per word & words    & tokens & Neighbors \\

    \midrule
    \multirow{2}{*}{\bpe}
        & \baseline & 0.157 & 0.524 & \textbf{1.146} & \textbf{0.53}\% & \textbf{6.00}\% & 2674 \\
        & \spl    & \textbf{0.179} & \textbf{0.527} & 1.165 & 0.98\% & 6.81\% & \textbf{2640} \\

    \midrule
    \multirow{2}{*}{\unigram}
        & \baseline & 0.151 & \textbf{0.505} & \textbf{1.162} & \textbf{0.56}\% & \textbf{9.42}\% & 2440 \\
        & \spl    & \textbf{0.171} & 0.485 & 1.176 & 1.00\% & 12.46\% & \textbf{2308} \\
         
    \bottomrule
    \end{tabular}
    \caption{Intrinsic benchmark results for Hebrew on a vocab size of 128K. The tokenizers were evaluated using the HeDC4 corpus. \textbf{Bold} values indicate better performance between \baseline\ and \spl.}
    \label{tab:intrinsic-hebrew-128}

\end{table*}

\subsection{Tokenizing with \spls}

With the list of reductions for each word length now sorted from highest to lowest scores, we apply it during inference for each word we encounter in a corpus or in task data, using a simple heuristic inspired by beam search to produce high-quality relinearizations.
We build a scored selection tree where the root is the full character sequence of the word scored at 1.0, and at every node select the $b$ highest-scoring applicable reductions according to the list, keeping the product of reduction scores so far as scores in the next level of nodes.
Once either the word length reaches the minimum of 3, or the depth of the tree reaches $d$, the reduction that started the highest-scoring path is selected and applied, and the process restarts with the new reduced word.\footnote{We set $b=d=3$.}

As described above, each reduction is encoded as a new composite character.
This transforms the word into a new representation as a sequence of this enriched alphabet, consisting of the original characters with the addition of the composite ones, which is then used as input for standard subword tokenization methods.
From the perspective of the tokenizer and the entire language model, both training and inference operate on this relinearized sequence represented by this new alphabet.
In generation mode, a character sequence over the new alphabet is decoded back into reductions, which are then applied sequentially to construct words.

\section{Intrinsic Evaluation}

To evaluate the effects of using \spl{} as a pre-processing step before tokenization, we trained multiple tokenizers on raw text from nonconcatenative languages and their \spl{}-treated counterparts.
We examined the performance of both the bottom-up \textbf{BPE} tokenization algorithm, which works by iteratively merging tokens based on corpus co-occurrence statistics, and the top-down \textbf{UnigramLM} approach, which starts with a very large vocabulary and iteratively removes from it tokens which contribute minimally to the corpus's likelihood.
We train over a wide range of vocabulary sizes in order to assess the utility of \spl{} in multiple scenarios: from multilingual models which can allocate roughly 1,000 tokens for a given language, to dedicated monolingual models with room for two orders of magnitude more tokens.
We selected three languages for our experiments: Hebrew and Arabic are Semitic languages featuring root-template morphology as discussed above, while Malay is an Austronesian language rich in \emph{circumfixes} and \emph{infixes}---morphemes which break either the stem or the affix when forming the composite inflection.
We computed the \spl{} operations for each language using the reductions map generated from the November 2023 Wikipedia dump of the respective language. 
The Wikipedia dump sizes were 1.9GB for Hebrew, 3.0GB for Arabic, and 0.4GB for Malay, and were downloaded using the Hugging Face \say{datasets} library.
We trained the tokenizers on the same Wikipedia dump used for \spl{} training, using Google's SentencePiece library with default settings, except for the tokenizer type (\unigram{} or \bpe{}) and vocabulary size (800, 1K, 2K, 10K, 32K, 64K, 128K).

\begin{table*}
    \centering
    \small
    \begin{tabular}{llrrrrrrr}
    \toprule

    Vocab & Type & Cognitive    & Rényi      & Tokens   & 4+ token & 1-char & Distinct \\
    size  &      & plausibility & efficiency & per word & words    & tokens & Neighbors \\

    \midrule
    \multirow{2}{*}{128K} 
        & \baseline & 0.157 & 0.524 & \textbf{1.146} & \textbf{0.53}\% & \textbf{6.00}\% & 2674 \\
        & \spl      & \textbf{0.179} & \textbf{0.527} & 1.165 & 0.98\% & 6.81\% & \textbf{2640} \\
    
    \midrule
    \multirow{2}{*}{64K} 
        & \baseline & 0.181 & 0.565 & \textbf{1.224} & \textbf{0.75}\% & \textbf{7.05}\% & 4272 \\
        & \spl      & \textbf{0.206} & \textbf{0.567} & 1.248 & 1.43\% & 8.35\% & \textbf{4188} \\

    \midrule
    \multirow{2}{*}{32K} 
        & \baseline & 0.201 & 0.610 & \textbf{1.336} & \textbf{1.12}\% & \textbf{8.76}\% & 5754 \\
        & \spl      & \textbf{0.223} & \textbf{0.612} & 1.365 & 2.04\% & 10.66\% & \textbf{5631} \\

    \midrule
    \multirow{2}{*}{10K} 
        & \baseline & 0.196 & \textbf{0.690} & \textbf{1.606} & \textbf{2.28}\% & \textbf{13.26}\% & 5652 \\
        & \spl      & \textbf{0.226} & 0.687 & 1.651 & 3.86\% & 16.56\% & \textbf{5555} \\

    \midrule
    \multirow{2}{*}{2K} 
        & \baseline & 0.149 & \textbf{0.760} & \textbf{2.137} & \textbf{7.44}\% & \textbf{25.90}\% & 1855 \\
        & \spl      & \textbf{0.207} & 0.756 & 2.270 & 11.57\% & 33.32\% & \textbf{1815} \\

    \midrule
    \multirow{2}{*}{1K} 
        & \baseline & 0.109 & \textbf{0.774} & \textbf{2.436} & \textbf{13.33}\% & \textbf{34.47}\% & 925 \\
        & \spl      & \textbf{0.184} & 0.763 & 2.713 & 22.82\% & 48.59\% & \textbf{895} \\

    \midrule
    \multirow{2}{*}{800} 
        & \baseline & 0.102 & \textbf{0.779} & \textbf{2.543} & \textbf{16.03}\% & \textbf{37.70}\% & 734 \\
        & \spl      & \textbf{0.182} & 0.762 & 2.890 & 28.70\% & 54.37\% & \textbf{705} \\
         
    \bottomrule
    \end{tabular}
    \caption{Intrinsic benchmark results for Hebrew using \bpe\ tokenizer with different vocabulary sizes. The tokenizers were evaluated using the HeDC4 corpus. \textbf{Bold} values indicate better performance between \baseline\ and \spl.}
    \label{tab:intrinsic-hebrew-bpe}

\end{table*}

We qualitatively examine the differences between the \baseline\ tokenizer and the \spl\-enhanced one.\footnote{All the examples in this paragraph are actual tokenizer outputs, comparing \baseline{} \bpe{} with \spl{} \bpe{}, both using a vocabulary size of 2K tokens.}
\autoref{fig:splintering} illustrates such differences: the \baseline\ \bpe\ tokenizer linearly segments the word \<l`bwd> `to work' into \say{\<l`>} and \say{\<bwd>}, thereby splitting the root \<`bd> across two tokens. 
In contrast, the non-linear \bpe\ + \spl\ tokenizer separates the word into two tokens: one for the template letters \say{3:\<w>,0:\<l>} and another for the root \say{\<`bd>}.
A similar pattern is seen in the noun \<.hy/swb> \emph{xi{\v s}uv} `computation', where the \baseline\ \bpe\ tokenizer linearly segments it into \say{\<.h>} and \say{\<y/swb>}, while \bpe\ + \spl\ produces a token for template letters \say{1:\<y>,3:\<w>} and another token to the root \say{\<.h/sb>}.
Likewise, for the definite adjective \<hp/sw.t> \emph{ha-pa{\v s}ut} `the simple.sg.masc', the \baseline\ tokenizer segments it into three tokens: \say{\<hp|>}, \say{\</s>} and \say{\<w.t>}, whereas \bpe\ + \spl\ separates it into \say{0:\<h>,3:\<w>} and the root \say{\<p/s.t>}.

For direct evaluation of the tokenizer vocabularies, independent of further language model architecture and training, we follow the analytical procedures collected in \citet{uzan-etal-2024-greed}, adding pairwise comparative measures from other sources as well.

\paragraph{Vocabulary overlap}
First, we validate that \spl{} provides models with vocabularies that are different enough from raw-text tokenizers.
Hypothetically, if many common words are learned in full as single tokens from raw text, there is no need for a special pre-processing step to account for an edge case.
However, in \autoref{fig:intersection} we show that this is not the case.
In all three languages, the maximal vocabulary size of 128K stays at an intersection level below 75\%, with the slope of added shared token rate declining to near constant.
Moreover, even if we assume a linear extrapolation rate, the intersection rate would only exceed 85\% at a vocabulary size of around 780K, which is exceptionally large and is not used even in SOTA English-dominated LLMs like GPT-4o. 
We note that we used a generous calculation for the intersection rate, as not all tokens in \spl-enhanced tokenizers can be directly compared to those in a regular tokenizer.
For instance, a token representing the reduction \say{0:\<l>,3:\<w>} cannot be linearly converted into a standard token. 
To make the comparison as permissive as possible, we applied the reductions within the token (e.g., converting \say{0:\<l>,3:\<w>} into \say{\<lw>}) and counted it towards the intersection if the resulting token existed in the \baseline\ tokenizer's vocabulary. 
As a result, the actual intersection percentage may be significantly lower than reported.
We conclude that \spl-enhanced tokenizers produce substantially different vocabularies at all stages of the vocabulary creation process.

\begin{table*}
    \centering
    \small
    \begin{tabular}{lllrrrrr}
    \toprule

    Language & Vocab & Type & Rényi      & Tokens   & 4+ token & 1-char & Distinct \\
             & size  &      & efficiency & per word & words    & tokens & Neighbors \\

    \midrule
    \multirow{4}{*}{Hebrew}
        & \multirow{2}{*}{128K} 
            & \baseline & 0.509 & \textbf{1.119} & \textbf{0.57}\% & \textbf{6.01}\% & 1463 \\
            & & \spl    & \textbf{0.511} & 1.134 & 0.92\% & 6.61\% & \textbf{1460} \\
        & \multirow{2}{*}{2K} 
            & \baseline & \textbf{0.779} & \textbf{2.149} & \textbf{9.22}\% & \textbf{26.34}\% & 1853 \\
            & & \spl    & 0.777 & 2.306 & 13.65\% & 33.81\% & \textbf{1805}\\

    \midrule
    \multirow{4}{*}{Arabic}
        & \multirow{2}{*}{128K} 
            & \baseline & 0.427 & \textbf{1.134} & \textbf{0.55}\% & \textbf{7.54}\% & \textbf{1444} \\
            & & \spl    & \textbf{0.430} & 1.158 & 1.07\% & 7.84\% & 1520 \\
        & \multirow{2}{*}{2K} 
            & \baseline & 0.736 & \textbf{2.117} & \textbf{11.25}\% & \textbf{29.37}\% & 1824 \\
            & & \spl    & \textbf{0.744} & 2.276 & 16.70\% & 37.91\% & \textbf{1784}\\

    \midrule
    \multirow{4}{*}{Malay}
        & \multirow{2}{*}{128K} 
            & \baseline & 0.471 & \textbf{1.088} & \textbf{0.55}\% & \textbf{4.45}\% & \textbf{337}\\
            & & \spl    & \textbf{0.479} & 1.135 & 1.56\% & 5.98\% & 354 \\
        & \multirow{2}{*}{2K} 
            & \baseline & 0.756 & \textbf{2.150} & \textbf{14.65}\% & \textbf{29.23}\% & 1215 \\
            & & \spl    & \textbf{0.770} & 2.724 & 28.79\% & 43.37\% & \textbf{1055} \\
         
    \bottomrule
    \end{tabular}
    \caption{Intrinsic benchmark results using \bpe\ tokenizer for Hebrew, Arabic and Malay on a vocab sizes of 2K and 128K. The tokenizers were evaluated using the Wikipedia corpus of their respective language. \textbf{Bold} values indicate better performance between \baseline\ and \spl.}
    \label{tab:intrinsic-languages}

\end{table*}

\paragraph{Cognitive plausibility} We use the metric introduced in \citet{beinborn-pinter-2023-analyzing} to measure the correlation of the tokenizer output with the response time and accuracy of human performance on a lexical decision task. 
This metric is based on the hypothesis that an effective tokenizer encounters difficulty with character sequences that are also challenging for humans, and vice versa. 
We use the Hebrew cognitive plausibility dataset~\citep[HeLP; ][]{stein2024help} to evaluate both the \bpe{} and \unigram{} tokenizers.
Each tokenizer was compared across seven vocabulary sizes, with the \baseline{} tokenizer evaluated against the \spl-enhanced tokenizer.
Following \citet{uzan-etal-2024-greed}, we report the average of the absolute value correlation scores across the four linguistic setups (word/nonword $\times$ accuracy/response time).
Higher scores mean better correlation with human performance.

As shown in \autoref{tab:intrinsic-hebrew-128} and \autoref{tab:intrinsic-hebrew-bpe}, \spl-enhanced tokenizers consistently correlate better with human lexical processing patterns, across all vocabulary sizes in both \bpe{} and \unigram.
These results suggest that downstream language models trained on \spl{} output would reach higher scores in morphological segmentation tasks, which we evaluate in \S\ref{sec:downstream}.
We note that, unlike the following metrics, cognitive plausibility does not focus on the tokenizer's effectiveness as a text compression tool, offering a different perspective.
Additional \unigram\ results provided in \autoref{app:unigram}.

\paragraph{Token distribution statistics}
We collected distributional data for the various tokenizers using the following corpora:
the HeDC4 corpus \citep{shalumov2023hero} was used in Hebrew experiments looking into vocabulary size and tokenizer type (\bpe\ vs \unigram), with a 10\% shuffled sample (seed = 42) taken from its original 45GB corpus.
For Hebrew's cross-linguistic comparison with Arabic and Malay, we used the respective November 2023 Wikipedia dump of each of the languages.\footnote{Prior to these experiments, we conducted a preliminary token distribution statistics evaluation of \spl{} on a small-sized pre-modern Hebrew corpus \citep{gershuni-pinter-2022-restoring}, 
yielding results consistent with those observed later on the other corpora. 
Details can be found in \autoref{app:pre_modern}.}
Based on these corpora, we report the Rényi efficiency score~\citep{zouhar-etal-2023-tokenization}, as well as several other surface statistics.
We report the Hebrew-specific results in \autoref{tab:intrinsic-hebrew-128} and \autoref{tab:intrinsic-hebrew-bpe}, and crosslinguistic results in \autoref{tab:intrinsic-languages}.
Rényi efficiency has been proposed as an indicator of downstream task performance, such as BLEU scores in machine translation.
This metric penalizes token distributions that are overly skewed toward either very high- and/or very low-frequency tokens.
However, a recent study~\citep{cognetta-etal-2024-two} suggests that this metric can be manipulated to produce higher scores while degrading actual performance.
This highlights the importance of using multiple indicators from different perspectives to make an informed assessment of a tokenizer’s potential impact on downstream tasks.

Rényi efficiency scores for Hebrew were consistent across both the HeDC4 and Wikipedia corpora, and in general show minimal differences between \spl\ tokenizers and \baseline\ tokenizers across most tokenization settings.
In \bpe, the \baseline\ tokenizers achieved slightly better results at lower vocabulary sizes ($\leq$2K), while in \unigram, the trend was reversed, with \spl\ tokenizers achieving slightly higher scores at lower vocabulary sizes ($\leq$2K), and the \baseline\ tokenizers achieving slightly higher scores at larger vocabulary sizes ($\geq$10K).
In Arabic and Malay, \spl's results were again very close to \baseline's, with \spl\ tokenizer scores slightly higher in both large and small vocabulary sizes.
The overall Rényi efficiency results suggest minimal impact on token distribution, with \spl\ sometimes slightly improving it and other times slightly reducing efficiency.

For further intrinsic evaluation, we examined three corpus-level indicators of tokenizer compression efficiency: the average number of tokens per word (also known as subword fertility), the percentage of words tokenized into four or more tokens, and the percentage of single-character tokens.
All three serve as indicators of compression efficiency, with lower values generally indicating better compression.

\begin{figure*}
    \centering
    \includegraphics[width=0.9\textwidth]{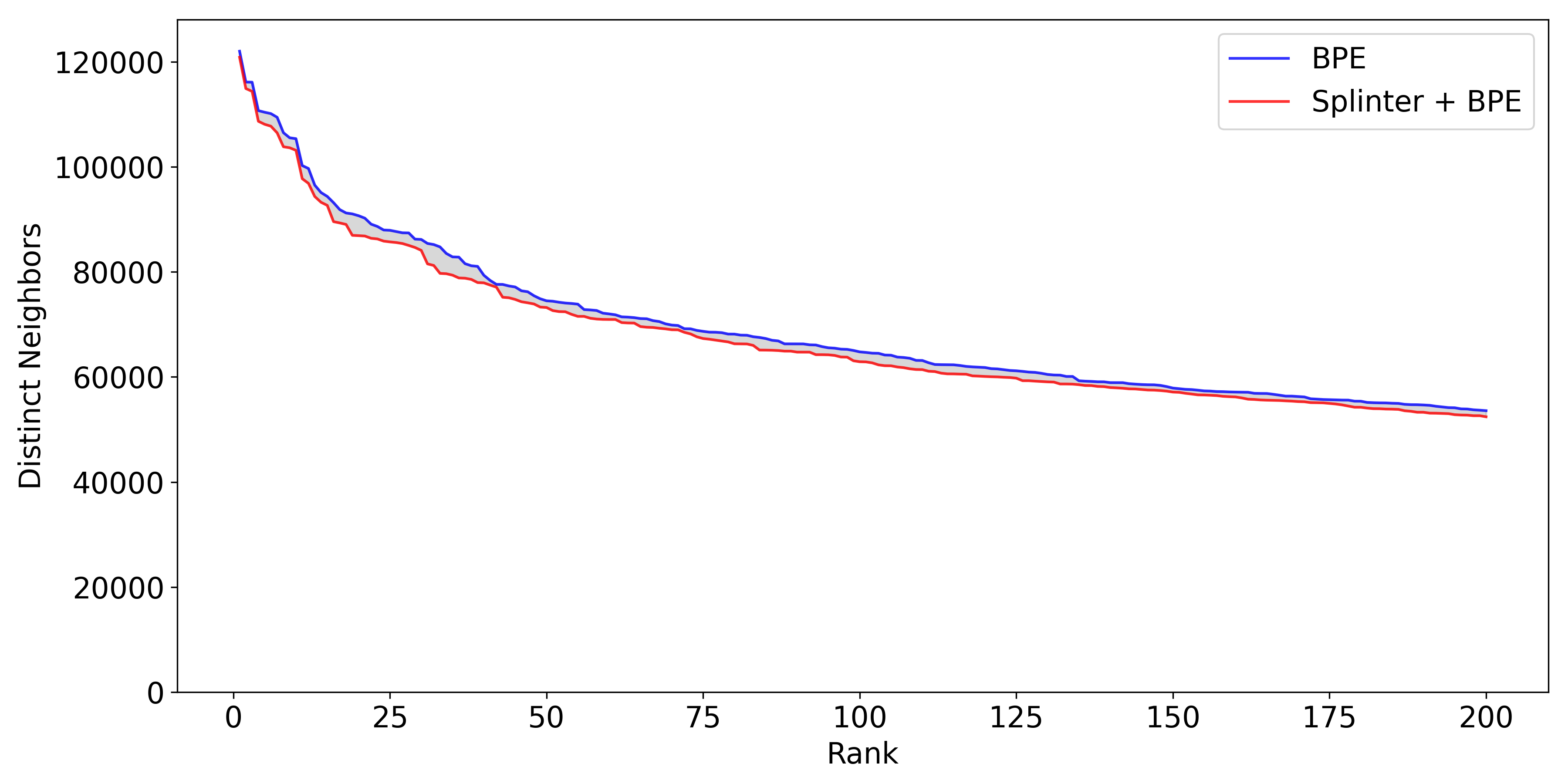}
    \caption{Distinct neighbor counts for top 200 tokens in \bpe\ and \spl\ + \bpe\ for a window of 2 on each side, vocabulary size 128K, HeDC4 corpus.}
    \label{fig:distinct-neighbors}
\end{figure*}

Across all corpora, tokenizer types, vocabulary sizes, and languages, the results consistently show that adding \spl\ to the tokenizer reduces its compression efficiency.
This result is important, as in generative LMs, for instance, less efficient compression requires more iterations to generate the same text, leading to higher computational costs.
Therefore, this trade-off should be considered when applying \spl\ in an LLM tokenizer.
That being said, \citet{schmidt-etal-2024-tokenization} found that tokenization should not be viewed principally from the compression perspective.
Improved compression does not always correlate with better downstream task performance, and in some cases, it may even degrade it.
This again emphasizes the importance of considering multiple perspectives when assessing tokenization quality.

\paragraph{Contextual coherence}
The next aspect we examine is the contextual coherence~\cite{yehezkel-pinter-2023-incorporating} of the tokens produced by each tokenizer, as measured by the number of distinct neighbors each token encounters within a window of $k$ tokens from each side (we choose $k=2$).
This measurement is motivated by the main downstream application scenario of vocabularies---contextualized embeddings in language models.
The fewer contexts a token appears in, the more likely a model is to learn a meaningful embedding for it over a corpus, as it offers better differentiation between token environments.
\autoref{fig:distinct-neighbors} displays the number of neighbors for the top 200 tokens in each tokenizer as ranked according to this quantity.
We present the average number of distinct neighbors across the vocabulary in Tables~\ref{tab:intrinsic-hebrew-128},~\ref{tab:intrinsic-hebrew-bpe}, and~\ref{tab:intrinsic-languages} as a measure of efficiency—not for text compression, but for downstream LM training. 
As shown in these tables, \spl-based tokenizers consistently produce fewer distinct neighbors than \baseline\ in Hebrew, regardless of vocabulary size, tokenization algorithm, or corpus. 
However, when evaluated on Wikipedia corpora, which were also used for training the tokenizers themselves, the differences were less pronounced.
\spl\ achieved a lower average number of distinct neighbors only at the 2K vocabulary size, while at 128K, its results were nearly identical to the baseline in Hebrew, and in Arabic and Malay, the baseline outperformed \spl.
These mixed results suggest that the impact of \spl\ on downstream tasks may depend on vocabulary size, with its advantages being more evident in smaller vocabularies dedicated to an NCL language, while in larger vocabulary sizes, the baseline tokenizer may perform better.

\section{Language Modeling with Splinters}
\label{sec:downstream}

\begin{table*}
    \centering
    % \small
    \begin{tabular}{lcccc}
        \toprule
        \textbf{Model} & \textbf{QA (F1)} & \textbf{QA (EM)} & \textbf{Syntax (LAS)} & \textbf{Seg (Acc)} \\
        \midrule
        \textit{Train} & \multicolumn{2}{c}{27K examples} & 82K words & 1.2M words \\
        \textit{Dev} & \multicolumn{2}{c}{1.5K examples} & 7K words & --- \\
        \textit{Test} & \multicolumn{2}{c}{1.5K examples} & 7K words & 307K words \\
        \midrule
        DictaBERT & 72.9 & 63.6 & \textbf{89.0} & 99.1 \\
        \spl & \textbf{74.4} & \textbf{65.4} & \textbf{89.0} & \textbf{99.3} \\
        \bottomrule
    \end{tabular}
    \caption{Performance comparison of the existing SOTA Hebrew BERT (DictaBERT) with our newly-pretrained \spl-based Hebrew BERT, across three downstream Hebrew NLP tasks.} %Dataset sizes are shown below model performance.}
    \label{tab:downstream-results}
\end{table*}

As noted above, standard subword tokenizers are suboptimal for nonconcatenative languages such as Hebrew and Arabic. Thus, we proceed now to evaluate \spl's impact on downstream NLP tasks for Hebrew. Specifically, we evaluate Prefix Segmentation and Syntactic Parsing~\citep{sade-etal-2018-hebrew,bareket-tsarfaty-2021-neural,zeldes-etal-2022-second}, and Question Answering~\citep{cohen-etal-2023-heq}.

We begin with the open \texttt{DictaBERT} BERT-base model~\cite{shmidman2023dictabert}, which delivers current SOTA-level performance on the aforementioned tasks~\cite{shmidman-etal-2024-mrl} in Hebrew.
We then pre-train a new BERT-base model using the same corpus and training parameters as \texttt{DictaBERT}, with the only modification being the \spl-processed tokenization.
We then fine-tune the new base model for the aforementioned tasks, and evaluate performance vis-a-vis fine-tuning the original \texttt{DictaBERT}.
We follow the same task parameters defined by \citet{shmidman-etal-2024-mrl} and \citet{shmidman2023dictabert} for the three tasks.
We present the results in \autoref{tab:downstream-results}, noting that these should be viewed as lower-bound estimates, since we retained the original \texttt{DictaBERT} training parameters and only tuned the \spl{} pre-tokenization step. With additional hyperparameter tuning of the language model itself, performance could potentially improve further.

\paragraph{Results}
Regarding both of the Question-Answer benchmarks, we find that the \spl-based model provides a substantial boost in performance.
We believe that this indicates that the \spl-based tokenization provides the model with a substantially stronger ability to process and understand the message of a Hebrew text, and thus to achieve superior performance on this high-level textual challenge.
On the sentence-level task of syntactic parsing, the performance of the \spl-based model is essentially the same as the existing \texttt{DictaBERT} model, indicating diminished advantage for manipulation at the character level.
However, for the nearly-saturated task of labeling prefix segmentation at the character level, \spl-based tokenization provides over 20\% reduction in errors, highlighting the effectiveness of the data-driven pattern-finding algorithm it employs.

\paragraph{Example 1: Question-Answer Task}
An illustrative example in which \spl's tokenization architecture allows it to succeed where \texttt{DictaBERT} fails is the following question from the QA corpus, regarding the date of a certain archaeological excavation.
The relevant part of the input text is comprised of the following three sentences (provided here in English translation):

\textcolor{purple}{\say{In 1913 he purchased the tract of land that covers most of the eastern slope of the City of David in Jerusalem, and persuaded the French Jewish archaeologist Raymond Weill to conduct excavations there.
This was done in response to the scandalous excavation of Montagu Parker in the City of David in 1911. Rothschild returned and financed another season of excavations, in 1923--1924, under Weill's direction.}}

The following question is then posed to the system:
\textcolor{blue}{\say{In what year were the first excavations conducted in the City of David?}}

\texttt{DictaBERT} incorrectly answers 1923 (as per the third sentence), while \spl{} correctly answers 1911 (as per the second sentence). \texttt{DictaBERT}'s failure to pull the correct answer from the second sentence likely stems from the fact that in the original Hebrew of that sentence, the words \say{excavation of Montague Parker} are phrased using the singular form \say{excavation}, with a suffixed possessive pronoun (\cjRL{xpyrtw}~\textit{xafirato}).
This word differs from the non-suffixed plural term \say{excavations} used in the question (\cjRL{xpyrwt}~\textit{xafirot}).
Crucially, the difference between the terms is not just the suffixed \cjRL{w}~\emph{o} at the end, but also a letter from the middle of the base term.
It is precisely discrepancies such as these that \spl{} aims to solve.
It stands to reason that \texttt{DictaBERT}'s incorrect answer stems from its inability to see the connection between these two terms; thus, it was unable to understand the relevance of the second sentence, and instead took its incorrect answer from the subsequent sentence, which includes an exact match for the term \say{excavations}.
In contrast, thanks to its new tokenization architecture, \spl{} recognizes the connection between the two disparate terms and correctly answers \say{1911}.

\paragraph{Example 2: Segmentation Task}
An analysis of the results on the 
prefix segmentation task serves to underscore the relative advantage of \spl{} over \texttt{DictaBERT} when it comes to concatenations of multiple proclitics.

The Hebrew language allows prepositions, conjunctions, relativizers, and definite articles to be prepended as proclitics.
In the simple case a single proclitic is prepended to a given word; such cases are well attested throughout the corpus, and such combinations are often tokens in and of themselves in the \texttt{DictaBERT} vocabulary.
However, Hebrew also allows concatenations of multiple proclitics, resulting in a string of 2--5 letters prepended to a single word.
This results in over 100 possible permutations of proclitic letters that can be prepended to a Hebrew word; however, any given instance of a multiletter prefix on a given word tends to be sparsely attested and thus not included as an independent token within the \texttt{DictaBERT} vocabulary. 

Indeed, a review of the segmentation task results reveals numerous cases of words with multi-letter prefixes where \texttt{DictaBERT} cautiously removes too few letters, anticipating the usual case of short and simple prefixes, while \spl{} correctly segments the full set of concatenated proclitics.

Actual examples of such from the test corpus include the following Hebrew words: \<w/stlmydyw>\ `and that his students.masc' (2-letter prefix); \</smp.hmymwt>\ `that are of carbohydrates' (2-letter prefix); \<w/smtr'yynyM> `and that are being interviewed.masc' (2-letter prefix); \<wm'wrK> `and from a length' (2-letter prefix); \<k/sb'.hd> `while in one.masc' (3-letter prefix).
In all of these cases, \texttt{DictaBERT}'s segmentation is one letter short, while \spl{} correctly identifies the boundary between the proclitics and the primary word.
Indeed, none of these words appear in \texttt{DictaBERT}'s vocabulary. 

In a few more unusual cases, we find the converse: \texttt{DictaBERT} removes \textbf{too many} letters for the prefix.
This happens with the words \<wmhy.s`yM> `and from bids' (2-letter prefix) and \</sb.sb.sw> `that protruded.pl' (1-letter prefix).
These two words, too, are not found in \texttt{DictaBERT}'s vocabulary, and in its struggle to parse the word, \texttt{DictaBERT} ends up incorrectly assuming a concatenation of an additional proclitic beyond the ones actually present.
\spl{} handles both correctly.

Our review of the results on the segmentation task thus highlights how \spl{}'s architecture allows it to successfully cope with the possibility of multi-proclitic concatenations even when the specific combination of prefix+word is seldom attested in naturally occurring Hebrew corpora, and not found within the vocabulary of the model.

\section{Conclusion}

In this work, we introduced \spl, a novel pre-processing method for subword tokenizers designed to improve downstream performance on nonconcatenative languages (NCLs). 
By applying an iterative reduction process, \spl\ restructures words in a way that better aligns with existing subword tokenizers. 
Our approach was designed with key constraints in mind: ensuring lossless transformation, compatibility with existing tokenization frameworks, and applicability across different vocabulary sizes and model types, whether for an NCL, a single-language LM, or a multilingual model based on English with a limited number of tokens allocated for nonconcatenative languages.

Through intrinsic evaluations, we demonstrated that \spl-enhanced tokenizers exhibit distinct vocabulary distributions compared to \baseline\ tokenizers. 
Cognitive plausibility metrics indicated that \spl\ improves alignment with human-like lexical processing, while our analysis of compression-related metrics revealed that \spl\ trades off slight reductions in compression efficiency for potentially better linguistic representation. 

Our downstream evaluation highlights \spl's impact, particularly on higher-level NLP tasks such as question answering and on character-critical tasks such as prefix segmentation.
The intermediate syntactic level appears to be less affected by the nonconcatenativity of Hebrew text.

In future work, we will extend the downstream evaluation to Arabic and other Semitic languages, as well as more languages exhibiting non-templatic nonconcatenative phenomena. 
Additionally, we plan to evaluate the performance of a large multilingual generative model on various tasks after incorporating \spl, examining its effectiveness in a broader linguistic context.

\section*{Limitations}

Rearranging text in order to improve representation of nonconcatenative features is a hard high-level problem, and we believe our work is a first step towards remedying this inherent mismatch between modeling and language data.
However, our concrete algorithm is still not universally-applicable, as shown by the difference between results on Semitic languages and on Malay.
Primarily, we attribute this to the property where each single-character pruning action must result in a valid corpus word, mostly limiting the scope of linearization to templatic morphology rather than also including infixation and circumfixation.

In addition, the increase in performance comes at the cost of less efficient token sequences, as found in our fertility analysis.
Overcoming this tradeoff is important for lowering the costs of running LLMs on low-resource languages, already lagging behind their high-resource counterparts.

\section*{Acknowledgments}
This research was supported in part by the Israel Science Foundation (grant No. 1166/23).
The work of the third author has been funded by the Israel Science Foundation (grant No. 2617/22) and by the European Union (ERC, MiDRASH, Project No. 101071829; Principal investigators: Nachum Dershowitz, Tel-Aviv University; Judith Olszowy-Schlanger, EPHE-PSL; Avi Shmidman, Bar-Ilan University, and Daniel Stoekl Ben Ezra, EPHE-PSL), for which we are grateful.
Views and opinions expressed are, however, those of the authors only and do not necessarily reflect those of the European Union or the European Research Council Executive Agency. Neither the European Union nor the granting authority can be held responsible for them.

We thank the reviewers for their comments.
We thank Craig Schmidt for comments on earlier versions.
Ibraheem Abo Shakra helped with implementation and evaluation of the Arabic models.

\bibliography{anthology,custom}
\bibliographystyle{acl_natbib}

\newpage

\appendix
\section{Alphabet Encoding Implementation Details}
\label{sec:alphabet-encoding-implementation-details}

To support the expanded set of characters introduced by \spl, which includes the original alphabet along with symbols representing single-letter reductions, we needed an alphabet capable of handling a large number of unique symbols.
In Hebrew, this expanded alphabet comprised 252 characters, while for Arabic and Malay, it grew to 400 and 460 characters, respectively.
Since the Unicode character sets for these languages do not offer enough distinct symbols, we mapped each new character to a unique Chinese character, leveraging the large character set available in the Chinese writing system. 
This was done as a workaround, as attempts to use the Unicode Private Use Areas (PUA) with the SentencePiece library were unsuccessful. 
This approach allowed the tokenization process to remain seamless from the language model’s perspective, as it processed the input as Chinese text, effectively encoding the original text.

\section{\spl\ Algorithm}
\label{app:algo}

Pseudocode for the \spl\ map creation algorithm is presented in \autoref{alg:splinter-train}.

\begin{algorithm*}
\caption{High-level algorithm for training \spl.}
% \small
% \footnotesize
% \tiny
\label{alg:splinter-train}
\begin{algorithmic}[1]
\Function{TrainSplinter}{\texttt{corpus}}

\State \texttt{freqMap $\gets$ GetWordFrequenciesFromCorpus(corpus)}

\State \texttt{reductions $\gets$ InitializeEmptyReductionsMap()}
\For{\texttt{length $\gets 4$ \textbf{to} maxWordLength}}
    \For{\texttt{word \textbf{in} freqMap[length]}}
        \For{\texttt{position \textbf{in} word}}
            \State \texttt{permutation $\gets$ GetWordWithoutLetter(word,position)}%InPosition(word,position)}
            \If{\texttt{permutation $\in$ freqMap[length - 1].keys}}
                 % \State \texttt{reduction $\gets$ format(\say{{position}:{word[position]}})}
                 \State \texttt{reduction $\gets$ Reduction(position,word[position])}
                 \State \texttt{frequency $\gets$ freqMap[length \,{-}\, 1][permutation]}
                 \State \texttt{reductions[length][reduction]\,{+}\,= frequency}
            \EndIf
        \EndFor
    \EndFor
\EndFor

\State \texttt{sortedReductions $\gets$ sortReductionsByScoreDesc(reductions)}

\State \texttt{selectedReductions $\gets$ InitializeEmptyReductionsMap()}
\For{\texttt{length $\gets 4$ \textbf{to} maxWordLength}}
    \For{\texttt{word \textbf{in} freqMap[length]}}
        \For{\texttt{reduction \textbf{in} sortedReductions[length]}}
            \State \texttt{\textbf{Extract} (position, letter) \textbf{from} reduction}
            \If{\texttt{word[position] == letter}}
                \State \texttt{permutation $\gets$ GetWordWithoutLetter(word,position)}%InPosition(word,position)}
                \If{\texttt{permutation $\in$ freqMap[length$ - 1$].keys}}
                     \State \texttt{frequency $\gets$ freqMap[length \,{$-$}\, 1][permutation]}
                     \State \texttt{selectedReductions[length][reduction]\,{+}\,= frequency}
                     \State \text{\textbf{Break}}
                \EndIf
            \EndIf
        \EndFor
    \EndFor
\EndFor

\State \texttt{\textbf{Return} selectedReductions}

\EndFunction
\end{algorithmic}
\end{algorithm*}

\section{\spl\ \unigram\ results}
\label{app:unigram}

Intrinsic benchmark results for Hebrew using \unigram\ tokenizer on HeDC4 corpus are presented in \autoref{tab:intrinsic-hebrew-unigram}.

\begin{table*}
    \centering
    \small
    \begin{tabular}{llrrrrrrr}
    \toprule

    Vocab & Type & Cognitive    & Rényi      & Tokens   & 4+ token & 1-char & Distinct \\
    size  &      & plausibility & efficiency & per word & words    & tokens & Neighbors \\

    \midrule
    \multirow{2}{*}{128K} 
        & \baseline & 0.151 & \textbf{0.505} & \textbf{1.162} & \textbf{1.00\%} & \textbf{9.42\%} & 2440 \\
        & \spl      & \textbf{0.171} & 0.485 & 1.176 & 0.56\% & 12.46\% & \textbf{2308} \\
    
    \midrule
    \multirow{2}{*}{64K} 
        & \baseline & 0.180 & \textbf{0.522} & \textbf{1.243} & \textbf{0.88\%} & \textbf{11.50\%} & 3907 \\
        & \spl      & \textbf{0.194} & 0.495 & 1.261 & 1.65\% & 16.21\% & \textbf{3640} \\

    \midrule
    \multirow{2}{*}{32K} 
        & \baseline & 0.191 & \textbf{0.526} & \textbf{1.363} & \textbf{1.46\%} & \textbf{14.41\%} & 5322 \\
        & \spl      & \textbf{0.208} & 0.496 & 1.391 & 2.74\% & 21.48\% & \textbf{4931} \\

    \midrule
    \multirow{2}{*}{10K} 
        & \baseline & 0.177 & \textbf{0.536} & \textbf{1.663} & \textbf{3.62\%} & \textbf{21.26\%} & 5267 \\
        & \spl      & \textbf{0.213} & 0.517 & 1.713 & 6.53\% & 31.63\% & \textbf{5064} \\

    \midrule
    \multirow{2}{*}{2K} 
        & \baseline & 0.136 & 0.590 & \textbf{2.250} & \textbf{11.65\%} & \textbf{33.40\%} & 1824 \\
        & \spl      & \textbf{0.196} & \textbf{0.618} & 2.424 & 20.70\% & 51.04\% & \textbf{1776} \\

    \midrule
    \multirow{2}{*}{1K} 
        & \baseline & 0.127 & 0.618 & \textbf{2.604} & \textbf{21.22\%} & \textbf{43.92\%} & 917 \\
        & \spl      & \textbf{0.185} & \textbf{0.659} & 2.877 & 32.48\% & 63.21\% & \textbf{881} \\

    \midrule
    \multirow{2}{*}{800} 
        & \baseline & 0.126 & 0.629 & \textbf{2.730} & \textbf{25.33\%} & \textbf{47.98\%} & 726 \\
        & \spl      & \textbf{0.177} & \textbf{0.673} & 3.060 & 37.71\% & 68.34\% & \textbf{693} \\
         
    \bottomrule
    \end{tabular}
    \caption{Intrinsic benchmark results for Hebrew using \unigram\ tokenizer with different vocabulary sizes. The tokenizers were evaluated using the HeDC4 corpus. \textbf{Bold} values indicate better performance between \baseline\ and \spl.}
    \label{tab:intrinsic-hebrew-unigram}

\end{table*}

\section{\spl\ pre\_modern results}
\label{app:pre_modern}

Intrinsic benchmark results for Hebrew using \bpe\ tokenizer on pre\_modern corpus (available at \url{https://github.com/elazarg/hebrew_diacritized/tree/master/pre_modern}), are presented in \autoref{tab:intrinsic-hebrew-pre-modern}.

\begin{table*}
    \centering
    \small
    \begin{tabular}{llrr}
    \toprule

    Vocab & Type    & Rényi          & Tokens         \\
    size  &         & efficiency     & per word       \\

    \midrule
    \multirow{2}{*}{128K} 
        & \baseline & 0.583          & \textbf{1.315} \\
        & \spl      & \textbf{0.586} & 1.349          \\
    
    \midrule
    \multirow{2}{*}{64K} 
        & \baseline & 0.608          & \textbf{1.406} \\
        & \spl      & \textbf{0.609} & 1.448          \\

    \midrule
    \multirow{2}{*}{32K} 
        & \baseline & \textbf{0.636} & \textbf{1.514} \\
        & \spl      & 0.629          & 1.575          \\

    \midrule
    \multirow{2}{*}{10K} 
        & \baseline & \textbf{0.677} & \textbf{1.744} \\
        & \spl      & 0.662          & 1.815          \\

    \midrule
    \multirow{2}{*}{2K} 
        & \baseline & \textbf{0.730} & \textbf{2.124} \\
        & \spl      & 0.711          & 2.266          \\

    \midrule
    \multirow{2}{*}{1K} 
        & \baseline & \textbf{0.751} & \textbf{2.347} \\
        & \spl      & 0.725          & 2.608          \\

    \midrule
    \multirow{2}{*}{800} 
        & \baseline & \textbf{0.756} & \textbf{2.427} \\
        & \spl      & 0.726          & 2.757          \\
         
    \bottomrule
    \end{tabular}
    \caption{Rényi efficiency and tokens per word results for Hebrew using \bpe\ tokenizer with different vocabulary sizes. The tokenizers were evaluated using the pre\_modern corpus. \textbf{Bold} values indicate better performance between \baseline\ and \spl.}
    \label{tab:intrinsic-hebrew-pre-modern}

\end{table*}

\end{document}